\pdfoutput=1

\documentclass[11pt]{article}

\usepackage[final]{acl}

\usepackage{times}
\usepackage{latexsym}

\usepackage[T1]{fontenc}

\usepackage[utf8]{inputenc}

\usepackage{microtype}

\usepackage{inconsolata}

\usepackage{times}
\usepackage{newtxtext,newtxmath}
\usepackage{graphicx}
\usepackage{latexsym}
\usepackage{booktabs}
\usepackage{multirow}
\usepackage{makecell}
\usepackage{tabularx}
\usepackage{amsmath}
\usepackage{booktabs} 
\usepackage{caption} 
\usepackage{booktabs}
\usepackage{siunitx}

\title{Navigating the Dual Facets: A Comprehensive Evaluation of Sequential Memory Editing in Large Language Models}

\author{Zihao Lin$^{\spadesuit}$ \quad Mohammad Beigi$^{\spadesuit}$ \quad Hongxuan Li$^{\heartsuit}$ \quad Yufan Zhou$^{\varheartsuit}$ \\ \textbf{Yuxiang Zhang}$^{\clubsuit}$ \quad \textbf{Qifan Wang}$^{\diamondsuit}$ \quad \textbf{Wenpeng Yin}$^{\vardiamondsuit}$ \quad \textbf{Lifu Huang}$^{\spadesuit}$\\
  $^{\spadesuit}$Virginia Tech \quad $^\heartsuit$Duke University \quad $^{\varheartsuit}$Adobe Research \\ $^\clubsuit$Waseda University \quad
  $^{\diamondsuit}$Meta AI \quad $^{\vardiamondsuit}$The Pennsylvania State University \\
  \texttt{\{zihaol,lifuh\}@vt.edu} 
  }

\begin{document}
{\makeatletter\acl@finalcopytrue
  \maketitle
}
\begin{abstract}
Memory Editing (ME) has emerged as an efficient method to modify erroneous facts or inject new facts into Large Language Models (LLMs). Two mainstream ME methods exist: parameter-modifying ME and parameter-preserving ME (integrating extra modules while preserving original parameters).  Regrettably, previous studies on ME evaluation have two critical limitations: (i) \emph{evaluating LLMs with single edit only}, 
neglecting the need for continuous editing, and (ii) \emph{evaluations focusing solely on basic factual triples}, overlooking broader LLM capabilities like logical reasoning and reading understanding. This study addresses these limitations with contributions threefold: (i) We explore how ME affects a wide range of fundamental capabilities of LLMs under sequential editing.  Experimental results reveal an intriguing phenomenon:  Most parameter-modifying ME consistently degrade performance across all tasks after a few sequential edits. In contrast, parameter-preserving ME effectively maintains LLMs' fundamental capabilities but struggles to accurately recall edited knowledge presented in a different format. (ii) We extend our evaluation to different editing settings, such as layers to edit, model size, instruction tuning, etc. Experimental findings indicate several strategies that can potentially mitigate the adverse effects of ME.
(iii) We further explain why parameter-modifying ME damages LLMs from three dimensions: parameter changes after editing,  language modeling capability, and the in-context learning capability. Our in-depth study advocates more careful use of ME in real-world scenarios.

\end{abstract}

\section{Introduction}

\begin{figure}[t] 
  \centering
  \includegraphics[scale=0.37]{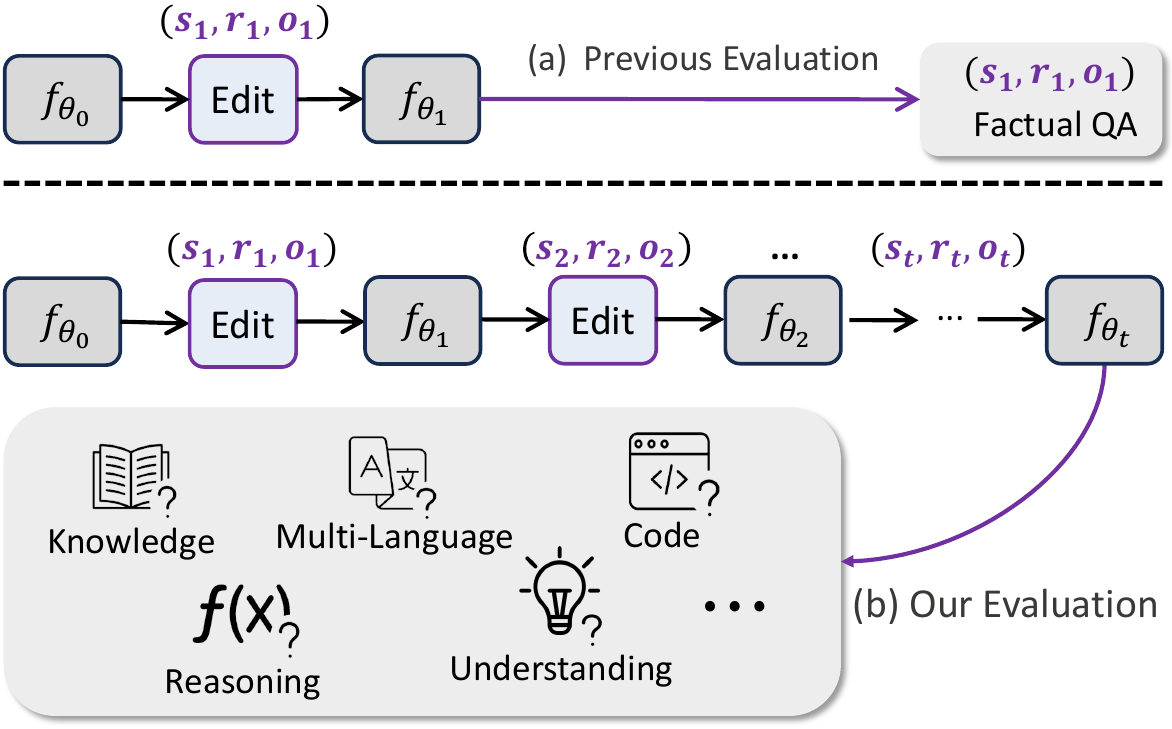}
  \caption{A comparison of two main limitations in previous memory editing evaluations. (a) shows the conventional method, assessing models after each edit, focused solely on the modified knowledge triples. (b) presents our approach, evaluating LLMs after a series of edits to assess their overall impact on various capabilities of LLMs, for a deeper insight into the enduring effects of memory editing.}
\vspace{-4mm}
  \label{illustration}
\end{figure}

Memory Editing (ME) was introduced as an effective method to correct erroneous facts or inject new knowledge into Large Language Models (LLMs).
Previous ME methods can be roughly divided into two categories: (1) parameter-modifying ME methods, for example, MEND \cite{mitchell2022fast}, ROME \cite{meng2022locating}, and MEMIT \cite{meng2022memit}, which directly modify a small number of parameters within the model, (2) parameter-preserving ME methods, such as GRACE \cite{hartvigsen2022aging} and MELO \cite{yu2023melo}, which integrate additional modules into the LLMs architecture without altering the original model parameters.

Although ME has shown much promise, previous studies evaluating and analyzing ME methods have two critical limitations, as depicted in Figure \ref{illustration}. 
First, they only consider the performance of LLMs after every single editing. However, in practice, LLMs usually need to be edited sequentially, i.e., sequential memory editing, which edits the same model multiple times to incorporate new knowledge continuously. Sequential memory editing is more important in real-world scenarios because new knowledge always appears over time. Second, prior research has predominantly concentrated on assessing ME's impact on factual knowledge. However, it is crucial to evaluate ME's influence on the broader capabilities of LLMs, such as logical reasoning, multilingual proficiency, code generation, and so on. 
Unfortunately, previous studies on evaluating and analyzing ME tend to overlook these broader aspects, hindering the popularity of ME methods in practical applications. 

To address these limitations, our study comprehensively evaluates the general capabilities of memory-edited LLMs in sequential editing scenarios. This evaluation involves four distinct ME methods, including three parameter-modifying ME methods - MEND \cite{mitchell2022fast}, ROME \cite{meng2022locating} and MEMIT \cite{meng2022memit}, and one parameter-preserving ME method - GRACE \cite{hartvigsen2022aging}. We leverage three different checkpoints of \texttt{LLaMA-2} \cite{touvron2023llama}, consisting of \texttt{LLaMA-2-7B}, \texttt{LLaMA-2-7B-Chat} and \texttt{LLaMA-2-13B} as base LLMs. The evaluation framework spans six core capabilities of LLMs: Professional Knowledge, Common Sense Knowledge, Logical Reasoning, Reading Understanding, Multilingual Proficiency, and Code Generation, based on eight downstream evaluation benchmarks.

The experimental findings reveal varied impacts of the parameter-modifying versus parameter-preserving ME methods on LLMs in sequential editing scenarios.
Specifically, all parameter-modifying ME methods systematically damage all fundamental capabilities of LLMs after a few sequential edits. On the contrary,  
the parameter-preserving ME method, GRACE \cite{hartvigsen2022aging}, effectively maintains the core capabilities of the model even after 100 sequential edits, without any noticeable degradation in the performance across various downstream tasks. However, models edited using GRACE exhibit limited \textit{generalization}, suggesting that the edited model struggles to recall the newly incorporated knowledge when it is presented in a different format. For example, if the edited knowledge is ``who is the CEO of Apple? Tim Cook'', the post-edited model cannot correctly answer the same question described differently - ``Who leads Apple as CEO?''

We then extend our analysis of parameter-modifying ME methods - the ROME and MEMIT, to more editing settings, including increasing the model size, instruction tuning, editing different layers, and the batch size of memory editing. Interestingly, experimental results indicate that larger models show more robustness on multilingual and code-generation tasks, while instruction tuning can alleviate the decline in knowledge QA tasks. Besides, editing deeper layers and increasing the batch size are also beneficial to maintain the general capabilities of LLMs. However, these strategies can not entirely overcome the observed performance decline. Our findings underscore the inherent complexity and challenges of applying ME in the sequential editing setting.

To explain how parameter-modifying ME methods damage the general capabilities of LLMs, we further analyze the post-edited models from three aspects: the changes in the model parameters, the language modeling capability, and the in-context learning capability. Experimental findings reveal that with each sequential edit, there is an increasing deviation in the model's parameters from those of the original model. This divergence is identified as the primary cause of noted performance damage. As a result of these parameter shifts, the language modeling capability of post-edited LLMs suffers a noticeable degradation after sequential edits. Interestingly, the post-edited LLMs can maintain the in-context learning capability when editing shallow and deep layers instead of middle layers. Our analysis provides insights into the understanding of parameter-modifying ME methods and sheds light on proposing new strategies to alleviate the damage or new ME methods in the future.

In summary, our study makes several pivotal contributions to the field:

\begin{itemize}
\vspace{-2mm}
    \item We pioneer a comprehensive evaluation of post-edited LLMs to assess their general capabilities in sequential memory editing scenarios. Our study uniquely covers both types of ME methods and examines their impacts across six core capabilities of LLMs, revealing distinct drawbacks. 
    \vspace{-2mm}
    \item Our comprehensive experiments suggest that instruction tuning, editing deeper layers, increasing model size, and increasing the batch size of memory editing are beneficial to mitigate the damage caused by the parameter-modifying ME methods, but cannot entirely overcome the adverse effect. 
    \vspace{-2mm}
    \item We analyze the damage of ME to LLMs in three dimensions: (1) parameter changes, (2) language modeling capability, and (3) in-context learning capability, which partially explains how memory editing influences LLMs, providing insights for the development of new ME methods and mitigation strategies. 
\end{itemize}

\section{Related Work}

\paragraph{Methods of Memory Editing}

\begin{figure}[t] 
  \centering
  \includegraphics[scale=0.38]{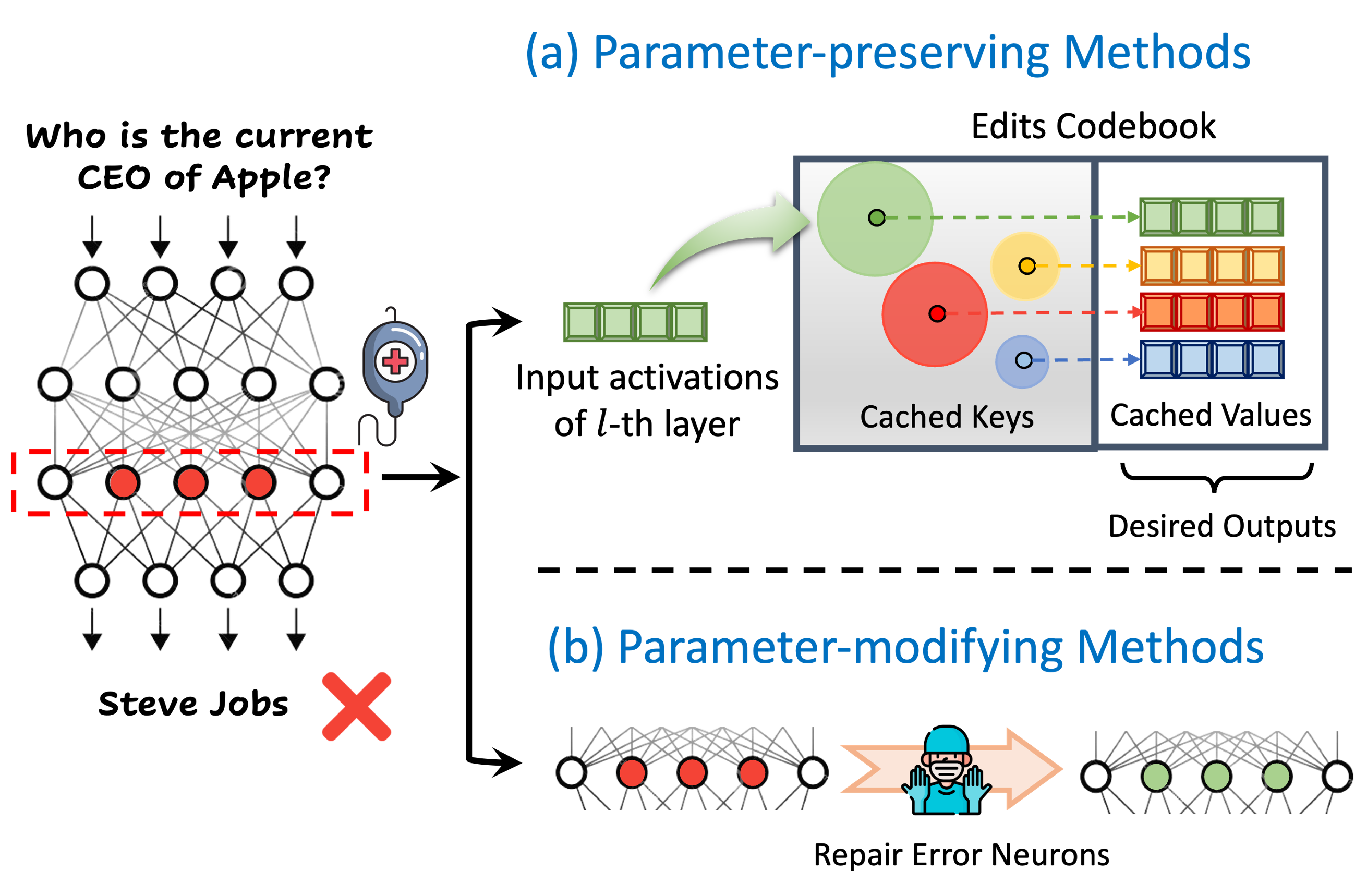}
  \caption{An overview of two categories of approaches for memory editing. We adopt GRACE~\cite{hartvigsen2022aging} as an example of the parameter-preserving ME method.
}
\vspace{-3mm}
  \label{fig:illustraition_editor}
\end{figure}

From the perspective of whether the model parameters are modified, previous ME methods can be divided into two categories: parameter-modifying ME methods 
and parameter-preserving ME methods 
\cite{yao2023editing}, as illustrated in Figure~\ref{fig:illustraition_editor}.
 KN \cite{dai2021knowledge}, an example of the parameter-modifying ME method, uses a knowledge attribution approach to identify and adjust relevant neurons in a Feed Forward Neural Network (FFN) layer. Similarly, ROME \cite{meng2022locating} and MEMIT \cite{meng2022memit} apply a Locate-Then-Edit strategy to inject new facts into LLMs. They first conduct causal analysis to pinpoint where the knowledge is stored in models and then edit the located parameters. Besides, meta-learning methods, for example, KE \cite{de2021editing} and MEND \cite{mitchell2022fast}, train a hypernetwork to estimate alterations or gradients of models' parameters for modification. Regarding the parameter-preserving ME methods, T-Patcher \cite{huang2023transformer} and CaliNET \cite{dong2022calibrating} introduce additional neurons into the FFN layer. GRACE \cite{hartvigsen2022aging} and MELO \cite{yu2023melo}, on the other hand, implement a discrete codebook to incorporate new knowledge. Besides, SERAC \cite{mitchell2022memory} proposes a \textit{counterfactual model} to handle the edited knowledge. Additionally, MemPrompt \cite{madaan2022memory}, and IKE \cite{zheng2023can} explore prompt-based or in-context learning strategies to update the knowledge of LLMs.

\paragraph{Evaluations and Analysis of Memory Editing}

Recently, in addition to exploring new ME methods, evaluation and analysis of ME methods have also drawn much attention. \citet{hase2023does} critically examines the limitations of causal tracing in determining the specific layers to be edited in LLMs. \citet{ju2023klob} contribute a novel benchmark for assessing knowledge localization methods in LLMs. The scope of evaluation also extends to more complex aspects of the robustness of ME. For instance, \citet{li2023unveiling} introduces a benchmark dataset, underscoring two significant areas of concern: Knowledge Conflict and Knowledge Distortion. Similarly, \citet{cohen2023evaluating} presents a dataset designed to evaluate ME methods in six challenging scenarios. In a related vein, \citet{li2023evaluating} proposes the DepEdit framework, which assesses ME methods by considering the interdependencies between a fact and its logical implications. 
Regrettably, prior studies predominantly evaluate post-edited models per edit rather than sequentially, focusing narrowly on basic factual triples. Despite \citet{pinter2023emptying}'s caution, there is a lack of experimental evidence, creating a gap in understanding. To address this, our study conducts comprehensive experiments, assessing the impact of ME methods on the general capabilities of LLMs in sequential editing scenarios. We provide detailed analyses explaining the performance decline across various tasks, offering insights for mitigating damage or proposing improved ME methods.

\section{Notation and Backgrounds}

Following \citet{meng2022locating}, we denote a fact as a triple form $(s, r, o)$, where $s$ represents a subject (e.g., Tim Cook), $r$ represents a relation (e.g., the CEO of) and $o$ represents an object (e.g., Apple). Given a model $f$ with parameter $\theta$, we have $f_{\theta}(s,r)=o$. Memory editing aims to directly edit a model's parameter: $ME(f_{\theta}) = f_{\theta'}$, to force the model to remember a new knowledge denoted as $(s, r, o')$, such that $f_{\theta'}(s,r)=o'$ without changing other irrelevant facts. In the sequential model editing problem, each edit is made to the model after the last edit. We denote $f_{\theta_0}$ as the original model, and $f_{\theta_{t-1}}$ as the result model after $t-1$ times edition. The $t$-th editing is $ME(f_{\theta_{t-1}}) = f_{\theta_{t}}$, satisfying $f_{\theta_{t}}(s_t, r_t) = o_t$, where $(s_t, r_t, o_t)$ is the $t$-th new knowledge.

\section{Experimental Settings}

\begin{table*}[hbpt]

\centering
\resizebox{0.8\textwidth}{!}{
\renewcommand{\arraystretch}{0.8}

\begin{tabular}{@{}llccccccccl@{}}
\toprule
Method                 & \multicolumn{1}{l}{Edit \#.} & \textbf{MMLU} & \textbf{MBPP} & \textbf{MATH} & \textbf{BBH} & \textbf{TyDiQA} & \textbf{C3} & \textbf{ComQA} & \textbf{AX-b}                & \multicolumn{1}{c}{\textit{Avg.}}    \\ \midrule
LLaMA             & \multicolumn{1}{l|}{0}         & $46.8$       & $18.2$       &  $3.4$        & $38.4$      & $26.8$         & $32.1$     & $49.6$         & \multicolumn{1}{c|}{$45.9$} & \multicolumn{1}{c}{$32.7$} \\ \midrule
\multicolumn{11}{c}{\textit{parameter-modifying ME methods}}                                                                                                                                                                                    \\ \midrule
\multirow{4}{*}{MEND}  & \multicolumn{1}{l|}{1}         & $47.2$        & $19.2$       & $3.26$        & $38.3$       & $26.4$          & $32.2$      & $50.6$          & \multicolumn{1}{c|}{$49.0$}  &          \multicolumn{1}{c}{$33.3$}                   \\
                       & \multicolumn{1}{l|}{10}             & $46.5$        & $0.0$         & $0.1$         &      $9.2$        & $18.7$          & $25.2$      & $44.8$          & \multicolumn{1}{c|}{$45.9$}  &        \multicolumn{1}{c}{$23.8$}   \\
                       & \multicolumn{1}{l|}{20}             & $35.2$        & $0.0$         & $0.0$         & $4.2$        & $9.8$           & $14.9$      & $11.0$          & \multicolumn{1}{c|}{$26.5$}  &       \multicolumn{1}{c}{$12.7$}              \\
                       & \multicolumn{1}{l|}{100}             & $25.4$        & $0.0$         & $0.0$         & $0.0$        & $0.0$           & $0.0$       & $0.0$           & \multicolumn{1}{c|}{$0.0$}   &     \multicolumn{1}{c}{$3.2$}                 \\ \midrule
\multirow{4}{*}{ROME}  & \multicolumn{1}{l|}{1}                   & $46.9$        & $17.6$        & $3.3$         & $38.4$       & $26.8$          & $32.0$      & $49.6$          & \multicolumn{1}{c|}{$45.5$}  &         \multicolumn{1}{c}{$32.5$}         \\
                       & \multicolumn{1}{l|}{10}                    & $46.6$        & $17.8$        & $3.3$         & $38.3$       & $27.0$          & $32.6$      & $50.2$          & \multicolumn{1}{c|}{$45.2$}  &  \multicolumn{1}{c}{$32.6$}                \\
                       & \multicolumn{1}{l|}{20}                   & $34.3$        & $18.4$        & $2.6$         & $33.8$       & $24.1$          & $28.9$      & $20.6$          & \multicolumn{1}{c|}{$51.5$}  &       \multicolumn{1}{c}{$26.8$}              \\
                       & \multicolumn{1}{l|}{100}                  & $25.5$        & $2.8$         & $1.0$         & $28.8$       & $8.0$           & $23.2$      & $19.0$          & \multicolumn{1}{c|}{$38.4$}  &   \multicolumn{1}{c}{$18.3$}               \\ \midrule
\multirow{4}{*}{MEMIT} & \multicolumn{1}{l|}{1}                & $46.7$        & $18.4$        & $3.4$         & $38.3$       & $26.8$          & $32.0$      & $50.6$          & \multicolumn{1}{c|}{$45.9$}  &       \multicolumn{1}{c}{$32.8$}         \\
                       & \multicolumn{1}{l|}{10}                 & $46.7$        & $16.6$        & $3.2$         & $37.8$       & $26.7$          & $32.9$      & $51.1$          & \multicolumn{1}{c|}{$45.4$}  &    \multicolumn{1}{c}{$32.6$}               \\
                       & \multicolumn{1}{l|}{20}               & $25.3$        & $16.6$        & $1.9$         & $32.4$       & $19.5$          & $15.5$      & $19.7$          & \multicolumn{1}{c|}{$31.2$}  &   \multicolumn{1}{c}{$20.3$}                \\
                       & \multicolumn{1}{l|}{100}              & $22.9$        & $0.0$         & $0.0$         & $0.0$        & $0.0$           & $0.0$       &        $0.49$  & \multicolumn{1}{c|}{$1.8$}   &    \multicolumn{1}{c}{$3.1$}       \\ \midrule
\multicolumn{11}{c}{\textit{parameter-preserving ME methods}}                                                                                                                                                                                   \\ \midrule
GRACE                  & \multicolumn{1}{l|}{100}               & $46.8$        &       $18.2$        &   $3.4$       & $38.4$       & $26.8$          &   $32.1$      & $49.6$          & \multicolumn{1}{c|}{$45.9$}        &        $32.7$               \\  \bottomrule
\end{tabular}

}
\caption{Evaluation of four ME methods on eight tasks under the sequential editing setting for the LLaMA-2-7B model. ``ComQA'' refers to the CommonsenseQA dataset. The scores for the MMLU, BBH, and TyDiQA datasets are the mean values derived from all respective subsets.}
\vspace{-3mm}
\label{tab:eval-results}
\end{table*}

\paragraph{Base LLMs.} We perform experiments on one of the most popular open-source large language models, LLaMA-2 \cite{touvron2023llama}, including three different checkpoints: \texttt{LLaMA-2-7B}, \texttt{LLaMA-2-7B-Chat}, and \texttt{LLaMA-2-13B}.

\paragraph{ME Methods.} In this study, we select ROME \cite{meng2022locating}, MEMIT \cite{meng2022memit}, and MEND \cite{mitchell2022fast} as representative examples of parameter-modifying ME methods, covering both Locate-Then-Edit methods (such as ROME and MEMIT) and hypernetwork methods (e.g., MEND). Regarding parameter-preserving ME methods, we opt for GRACE \cite{hartvigsen2022aging}, a state-of-the-art method, as our chosen method. Considering that MELO \cite{yu2023melo} is built upon the same foundational framework and employs the same constraint method as GRACE, we decide to solely focus on GRACE. Furthermore, in-context learning approaches are excluded from our study, given that they do not modify parameters or even add new modules into LLMs.

\paragraph{Datasets.} We randomly select 100 samples from the ZsRE \cite{levy2017zero} as the editing dataset. To fully evaluate the fundamental capabilities of LLMs, we consider six core aspects: Professional Knowledge, Common Sense Knowledge, Logical Reasoning, Reading Understanding, Multilingual Proficiency, and Code Generation. Our evaluation framework consists of eight main benchmarks: MMLU \cite{hendrycks2020measuring}, BBH \cite{ghazal2013bigbench}, MATH \cite{hendrycks2021measuring}, SuperGLUE-AX-b \cite{wang2019superglue}, CommonsenseQA \cite{talmor2018commonsenseqa}, C3 \cite{sun2020investigating}, TydiQA \cite{clark2020tydi}, and MBPP \cite{austin2021program}. Details of the experimental settings and metrics corresponding to each dataset are shown in Appendix \ref{sec:appendix-evaluationg-datasets}.

\paragraph{Evaluation Metrics.} To evaluate whether the post-edited model can successfully answer questions about the new knowledge, we utilize \textit{reliability}, which checks if the edited model successfully remembers the added knowledge, and \textit{generalization}, which checks if the edited model recalls the new knowledge described in different formats. In our experiments, we only use one different format for each knowledge to calculate \textit{generalization}. 
In sequential editing scenarios, we define the \textit{individual reliability} and \textit{individual generalization} score to specifically assess the model's accuracy on the latest edit made in the most recent iteration. These scores evaluate how effectively the model integrates new information after each editing cycle. Conversely, \textit{sequential reliability} and \textit{sequential generalization} provide broader evaluations of the model's performance, considering the knowledge edits from all previous iterations, not just the recent ones. 

\section{Evaluations of ME on LLMs}
\label{sec:impact-of-sme-on-llm}

In this section, we explore the impact of the two types of ME methods on LLMs in sequential editing scenarios, aiming to quantify their damage to the general capabilities of LLMs.

\subsection{Evaluation of Parameter-Modifying ME Methods}

The evaluation results of the post-edited models on eight datasets are shown in Table \ref{tab:eval-results}.
Following the initial edit, all the ME methods maintain performance levels comparable to the baseline model on eight benchmarks. However, after 10 sequential edits, notable performance degradation is observed with the MEND method, particularly in benchmarks such as MBPP, MATH, TyDiQA, and C3. This decline contrasts with other methods that show relatively stable performance.
After 20 edits, a significant performance drop is evident in all three parameter-modifying ME methods across all evaluation datasets. 
After 100 sequential edits, the MEMIT and MEND fail in all tasks with nearly zero scores except the MMLU dataset. Note that, as described in Appendix \ref{sec:appendix-evaluationg-datasets}, each data instance in the MMLU dataset comprises a question and four possible answers, thus a random choice score should be around 25\% which is similar to the evaluation scores of all parameter-modifying ME methods after 100 sequential edits, indicating that the post-edited LLMs fail to answer all questions in the MMLU dataset. All these results highlight the systematic hurt of the parameter-modifying ME methods on LLMs
in sequential editing scenarios.

We report the individual and sequential scores of \textit{reliability} and \textit{generalization} in Table \ref{tab:reliability-and-generalization}. The decline of the sequential \textit{reliability} and \textit{generalization} indicates that in sequential editing scenarios, post-edited models, edited by parameter-modifying ME methods, forget previously edited knowledge after several edits. Besides, the individual \textit{reliability} and \textit{generalization} of the ROME and MEMIT methods remain similar as the number of edits increases, while the MEND method has a significant decline, indicating that in sequential editing scenarios, the MEND method cannot successfully add new knowledge into LLMs after several edits.

\subsection{Evaluation of Parameter-Preserving ME Method}
\label{sec:trade-off-grace-threshold}
The parameter-preserving ME method, GRACE, introduces an additional codebook to store edited knowledge. As described in Appendix \ref{appendix:grace}, it applies a threshold to control whether the input information uses the stored knowledge. In the experiments in Table \ref{tab:eval-results} and Table \ref{tab:reliability-and-generalization}, we set 1 as the value of the threshold. It is shown that such a small threshold helps maintain the broad capabilities of LLMs with no noticeable decline in the performance on all downstream tasks. However, it also restricts the post-edited model from correctly answering the question about the edited knowledge described in a different format. This results in a low score of \textit{generalization}, as illustrated in Table \ref{tab:reliability-and-generalization}. We claim that a larger threshold increases the \textit{generalization} but fails to preserve the core capabilities of LLMs. We discuss the influences of the threshold in Appendix \ref{appendix:trade-off-of-the-threshold-in-grace}.

\begin{table}[t]

\centering

\small

\begin{tabular}{@{}ll|cc|cc@{}}
\toprule
\multicolumn{1}{c}{} & \multicolumn{1}{c}{} & \multicolumn{2}{c|}{\textit{Sequential Score}} & \multicolumn{2}{c}{\textit{Individual Score}} \\
\cmidrule(lr){3-6}
Method                 & \multicolumn{1}{l|}{Edit \#.} & \textbf{Rel.} & \textbf{Gen.} & \textbf{Rel.} & \textbf{Gen.}\\

\midrule
\multicolumn{6}{c}{\textit{parameter-modifying ME methods}}   \\ \midrule
\multirow{4}{*}{MEND}  & \multicolumn{1}{l|}{1}         & $80$        & \multicolumn{1}{c|}{$80$}       & $80$        & $80$   \\
                       & \multicolumn{1}{l|}{10}             & $79.3$        & \multicolumn{1}{c|}{$74.8$}        & $86.8$        &      $87.7$  \\
                       & \multicolumn{1}{l|}{20}             & $39.1$        & \multicolumn{1}{c|}{$44.1$}         & $67.2$         & $68.1$     \\
                       
                       & \multicolumn{1}{l|}{100}             & $0$        & \multicolumn{1}{c|}{$0$}         & $13.6$        & $13.9$    \\ \midrule
\multirow{4}{*}{ROME}  & \multicolumn{1}{l|}{1}                   & $80$        & \multicolumn{1}{c|}{$80$}        & $80$         & $80$   \\
                       & \multicolumn{1}{l|}{10}                    & $66.7$        & \multicolumn{1}{c|}{$69.7$}        & $93$         & $87.3$   \\
                       & \multicolumn{1}{l|}{20}                   & $53.3$        & \multicolumn{1}{c|}{$52.4$}        & $90.3$         & $85.7$   \\
                       
                       & \multicolumn{1}{l|}{100}                  & $52.3$        & \multicolumn{1}{c|}{$49.5$}         & $93.3$         & $90.4$ \\ \midrule
\multirow{4}{*}{MEMIT} & \multicolumn{1}{l|}{1}                & $80$        & \multicolumn{1}{c|}{$80$}        & $80$         & $80$    \\
                       & \multicolumn{1}{l|}{10}                 & $87$        & \multicolumn{1}{c|}{$87$}        & $86.4$         & $83.2$   \\
                       & \multicolumn{1}{l|}{20}               & $22.4$        & \multicolumn{1}{c|}{$25.3$}        & $88.3$         & $88.1$                 \\
                       & \multicolumn{1}{l|}{100}              & $0.07$        & \multicolumn{1}{c|}{$0.06$}         & $87.7$         & $85.4$ \\ \midrule
\multicolumn{6}{c}{\textit{parameter-preserving ME methods}}       \\ \midrule
GRACE                  & \multicolumn{1}{l|}{100}               & $99.8$        &       \multicolumn{1}{c|}{$30.2$}        &   $99.8$       & $30.2$   \\  \bottomrule
\end{tabular}
\caption{The individual and sequential scores of \textit{reliability}, denoted as \textbf{Rel.} and \textit{generalization}, denoted as \textbf{Gen}. We evaluate the scores on the editing dataset. }
\vspace{-4mm}
\label{tab:reliability-and-generalization}
\end{table}

\section{Impact of Different Editing Settings}
\label{sec:impact-of-different-editing-settings}

This section is dedicated to analyzing the influences of ME in different editing settings. We focus on four aspects: model size, instruction tuning, layers to edit, and the batch size of memory editing.

\paragraph{Model Size.} 

\begin{figure}[ht]
  \centering
  \includegraphics[width=\linewidth]{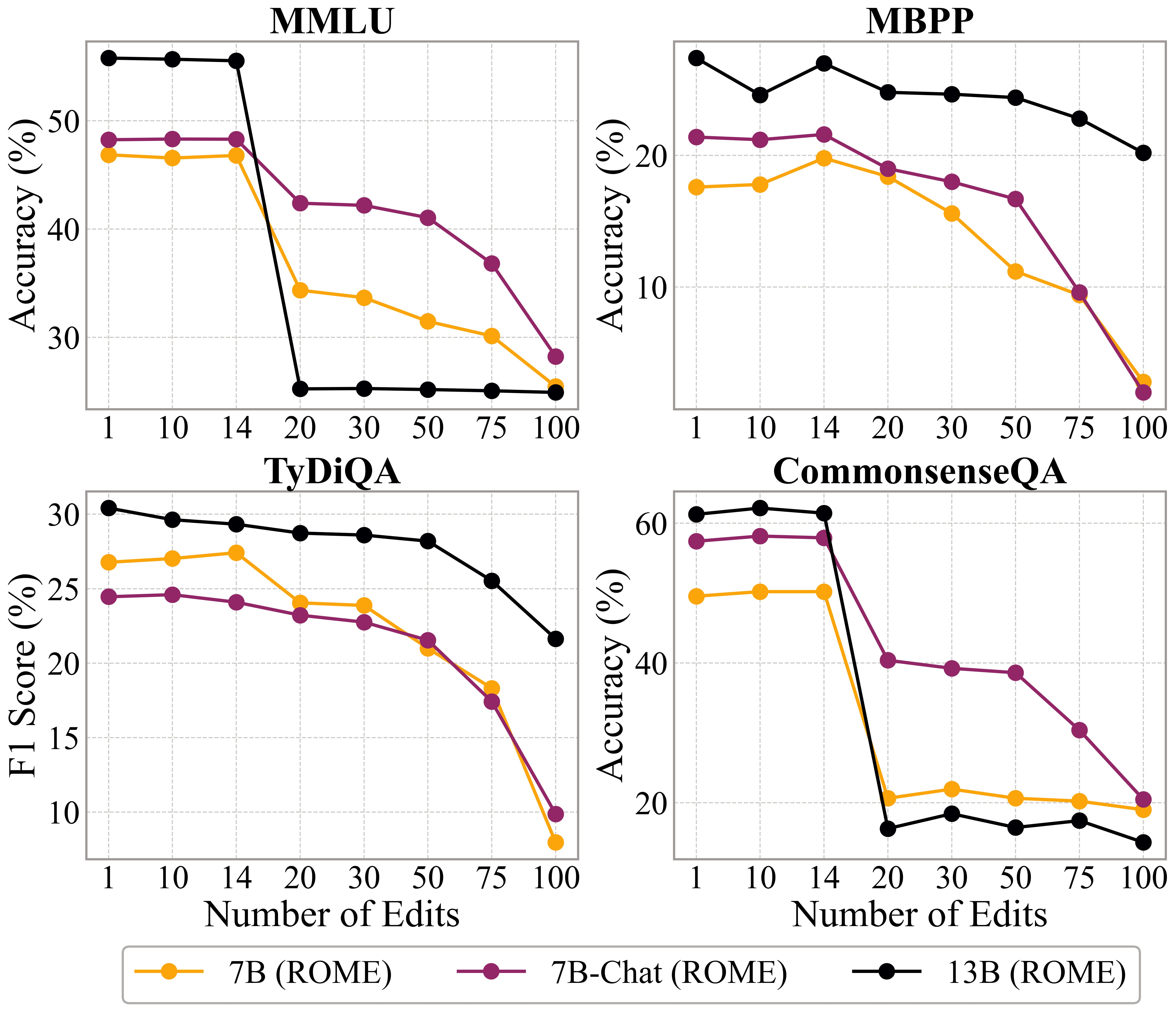}
  \caption{Evaluation of three different checkpoints of LLaMA-2-7B on four datasets. We apply ROME as the ME method. 
  }
  \vspace{-4mm}
  \label{fig:compare-checkpoints}
\end{figure}

Figure \ref{fig:compare-checkpoints} illustrates that all model checkpoints edited by the ROME method, regardless of their size, show performance degradation that correlates with the number of sequential edits. Interestingly, an increase in model size appears to have a protective effect, particularly in multilingual understanding and code generation domains, as shown in the TyDiQA and MBPP datasets. However, this protection does not extend to all areas. The post-edited LLMs with different sizes of 7B and 13B suffer the same decline trend on knowledge question-answering tasks, e.g., the MMLU and CommonsenseQA datasets. We conjecture the reason as: edited knowledge triples and the concerned knowledge in the MMLU and CommonsenseQA datasets are stored closer in the model's parameters, compared to the concerned knowledge of multi-lingual understanding or code generation. As models scale up, a more precise separation between the edited knowledge and concerned knowledge of code generation and multi-lingual understanding tasks emerges, potentially allowing for less disruptive memory editing. We leave the proof of these hypotheses as future work.

\paragraph{Instruction Tuning.} Compared with \texttt{LLaMA-2-7B}, \texttt{LLaMA-2-7B-Chat} is further instruction tuned to generate more natural conversational responses. The implementation of instruction tuning, particularly in the \texttt{LLaMA-2-7B-Chat} model, provides insightful observations. As shown in Figure \ref{fig:compare-checkpoints}, despite the overall performance degradation trend, instruction tuning appears to impart a degree of robustness, as evidenced by the enhanced stability across MMLU and CommonsenseQA. This finding suggests that instruction tuning might play a role in safeguarding model capabilities against the detrimental effects of memory editing, especially for knowledge question-answering tasks, although it does not entirely prevent performance losses. However, instruction tuning does not help mitigate the damage to code generation and multi-lingual understanding tasks. 
The impact of instruction tuning on memory editing suggests an intriguing area for further investigation, especially regarding how it influences the model's capability to integrate and handle edited information. 

\begin{figure}[ht]
  \centering
  \includegraphics[width=\linewidth]{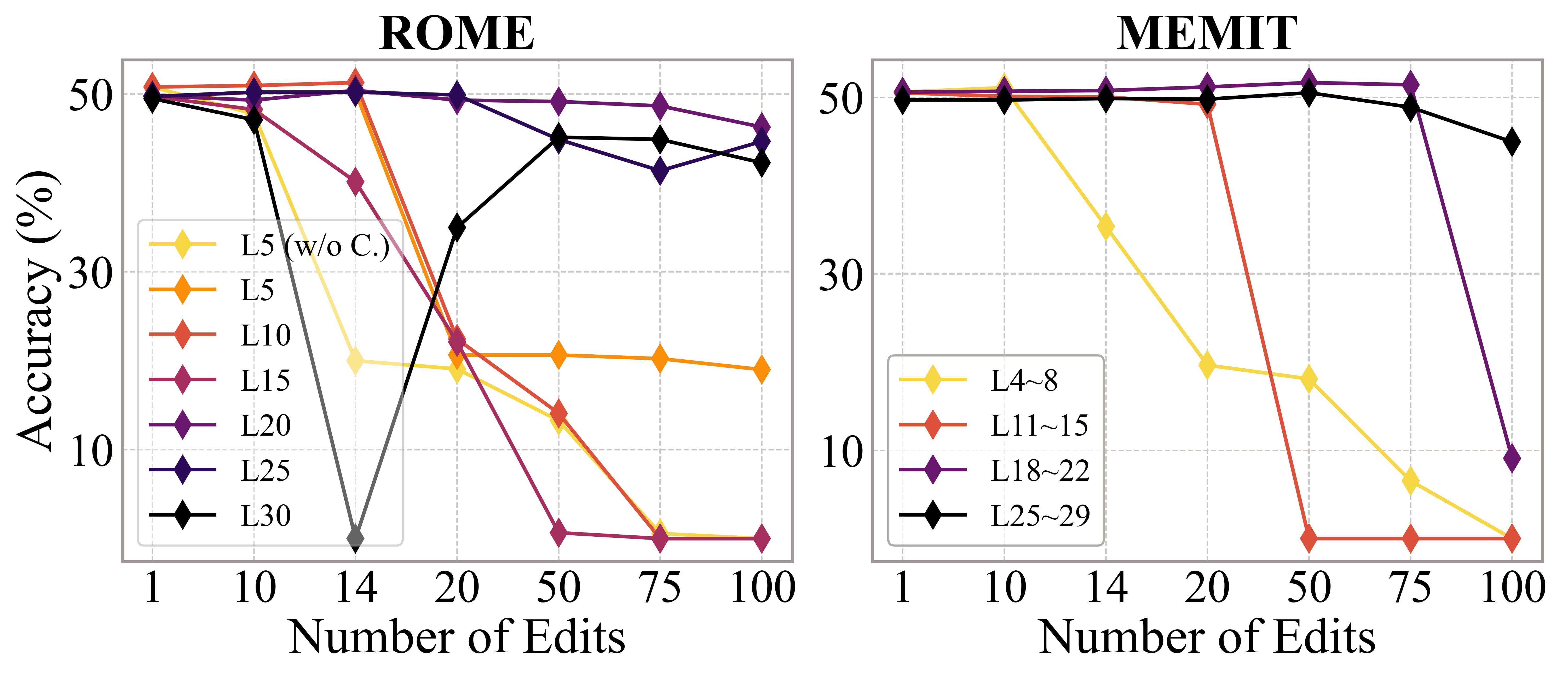}
  \caption{The performance of the LLaMA-2-7B model on the CommonsenseQA dataset. LX represents editing the X-th layer of the model, while LX\textasciitilde Y represents editing layers between the X-th and the Y-th layer. }
  \vspace{-4mm}
  \label{Fig:compare-edit-layer}
\end{figure}

\paragraph{Layers to Edit.} Inspired by \cite{hase2023does}, we investigate the effects of editing different layers in LLMs using the ROME and MEMIT methods. Figure \ref{Fig:compare-edit-layer} shows a noticeable trend: editing layers closer to the output (deeper layers) results in a marginal decrease in performance while editing shallower layers leads to significant performance degradation. Specifically, when editing the 20th layer of the \texttt{LLaMA-2-7B }model using ROME, the model's performance on CommonsenseQA after 100 editing iterations stands at 46.27\%\footnote{An intriguing observation emerges when we edit the 30th layer using ROME, which is explained in Appendix \ref{sec:Explanation-of-editing-deeper--layers}. }. However, editing shallower layers, such as the 5th, 10th, and 15th layers, severely impacts the model's performance, leading to significant deterioration after just 20 edits.  Similarly, with MEMIT, editing layers 25 through 29 leads to a performance decrease of just 9.6\% from the post-first-edit outcomes. 
These results indicate that the choice of layers for editing in LLMs significantly impacts their general capabilities, with deeper layers showing more resilience to the editing process than shallower ones. We also edit different layers using GRACE, whose results are shown in Appendix \ref{appendix:layers-to-edit-in-GRACE-method}, suggesting a similar conclusion as both ROME and MEMIT.

\begin{figure}[ht]
  \centering
  \includegraphics[width=0.8\linewidth]{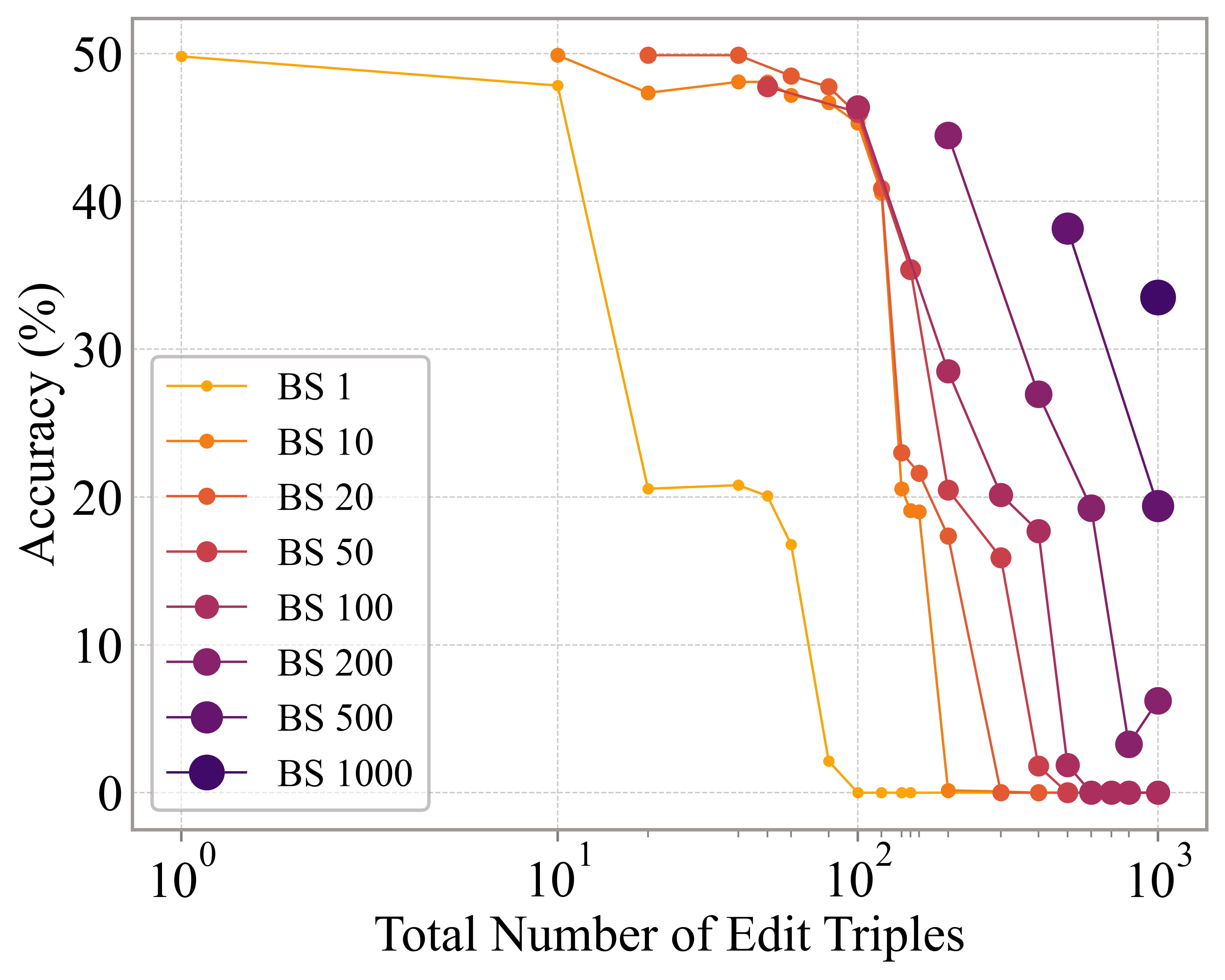}
  \vspace{-3mm}
  \caption{The performance of LLaMA-2-7B on CommonsenseQA, utilizing MEMIT as the editing method with different batch sizes for memory editing. The x-axis denotes the total number of edit triples. For example, for the line of batch size 100, the first data point of this line lies in the total number of edit triples 100, which only edits the model once. BS denotes batch size.}
  \vspace{-4mm}
  \label{Fig:compare-edit-batch-size}
\end{figure}

\paragraph{Batch Size of ME.} In line with \citet{meng2022memit}, we conduct experiments to test the influence of varying batch sizes of memory editing. Utilizing MEMIT to edit \texttt{LLaMA-2-7B}, we alter the batch size from 1 to 1000. As shown in Figure~\ref{Fig:compare-edit-batch-size}, 
with the same number of edit triples, increasing the batch size means reducing the number of editing times, which turns out to be beneficial in mitigating the damage of ME to LLMs.

\section{Interpreting Disruptions in LLMs Caused by Memory Editing}
\label{sec:interpreting-memory-editing-disruptions-in-llms}
To interpret the damage caused by parameter-modifying ME methods, our investigation is structured around three pivotal aspects: (i) the model parameter changes after being sequential edited, (ii) the impact on the language modeling capability of LLMs, and (iii) the in-context learning capacity. This multifaceted exploration is designed to provide a holistic understanding of how memory editing affects LLMs.

\begin{table*}[t]
\centering
\resizebox{0.8\textwidth}{!}{%
\renewcommand{\arraystretch}{1.2}
\begin{tabular}{
  S[table-format=2.0]
  S[table-format=5.2]
  S[table-format=5.2]
  S[table-format=5.2]
  S[table-format=5.2]
  S[table-format=5.2]
  S[table-format=5.2]
  S[table-format=5.2]
  S[table-format=5.2]
}
\toprule
\multicolumn{1}{c}{} & \multicolumn{8}{c}{Number of Edits} \\
\cmidrule(lr){2-9}
{Edit Layer} & {1} & {10} & {14} & {20} & {30} & {50} & {75} & {100} \\
\midrule
5   & 7.63   & 7.65   & 7.61   & 14.29  & 14.29  & 13.89  & 12.90   & 14.04   \\
10  & 7.15   & 7.32   & 7.38   & 28.61  & 45.23  & 81.08  & {/}     & {/}     \\
15  & 7.61   & 7.48   & 81.24  & 50.09  & 21.25  & 28.48  & 29634.93& 17220.91\\
20  & 7.61   & 7.75   & 7.69   & 8.12   & 8.09   & 9.48   & 10.67   & 11.15   \\
25  & 7.63   & 7.61   & 7.73   & 8.81   & 15.77  & 18.35  & 31.26   & 9830.27 \\
30  & 7.65   & 810.04 & 2477.53& 603.46 & 49.09  & 78.39  & 1018.46 & 1444.29 \\
\bottomrule
\end{tabular}
}
\caption{Perplexity scores when editing different layers with varying numbers of edits. We use LLaMA-2-7B as the base LLM to edit. The result ``/'' means that the edited model fails to generate any response. 
}
\vspace{-4mm}
\label{tab:perplexity-results}
\end{table*}

\subsection{Parameter Changes after Memory Editing}
\label{sec:parameter-changes-after-memory-editing}
In this section, we investigate the changes between the parameters of LLMs before and after sequential memory editing. We apply ROME as the ME method and \texttt{LLaMA-2-7B} as the base model. 
We use the Pearson product-moment correlation coefficient (\(R\)) to measure the similarities between the parameters of the original and edited layers within the model. The correlation coefficient matrix, \(R\), ranges from -1 to 1. An \(R\) value of 1 indicates a perfect positive linear correlation, implying that the parameters in both the original and edited layers are identical. Conversely, a value of -1 indicates a perfect negative correlation, while a value of 0 suggests no similarity between the parameters."

\begin{figure}[ht]
  \centering
  \includegraphics[width=0.9\linewidth]{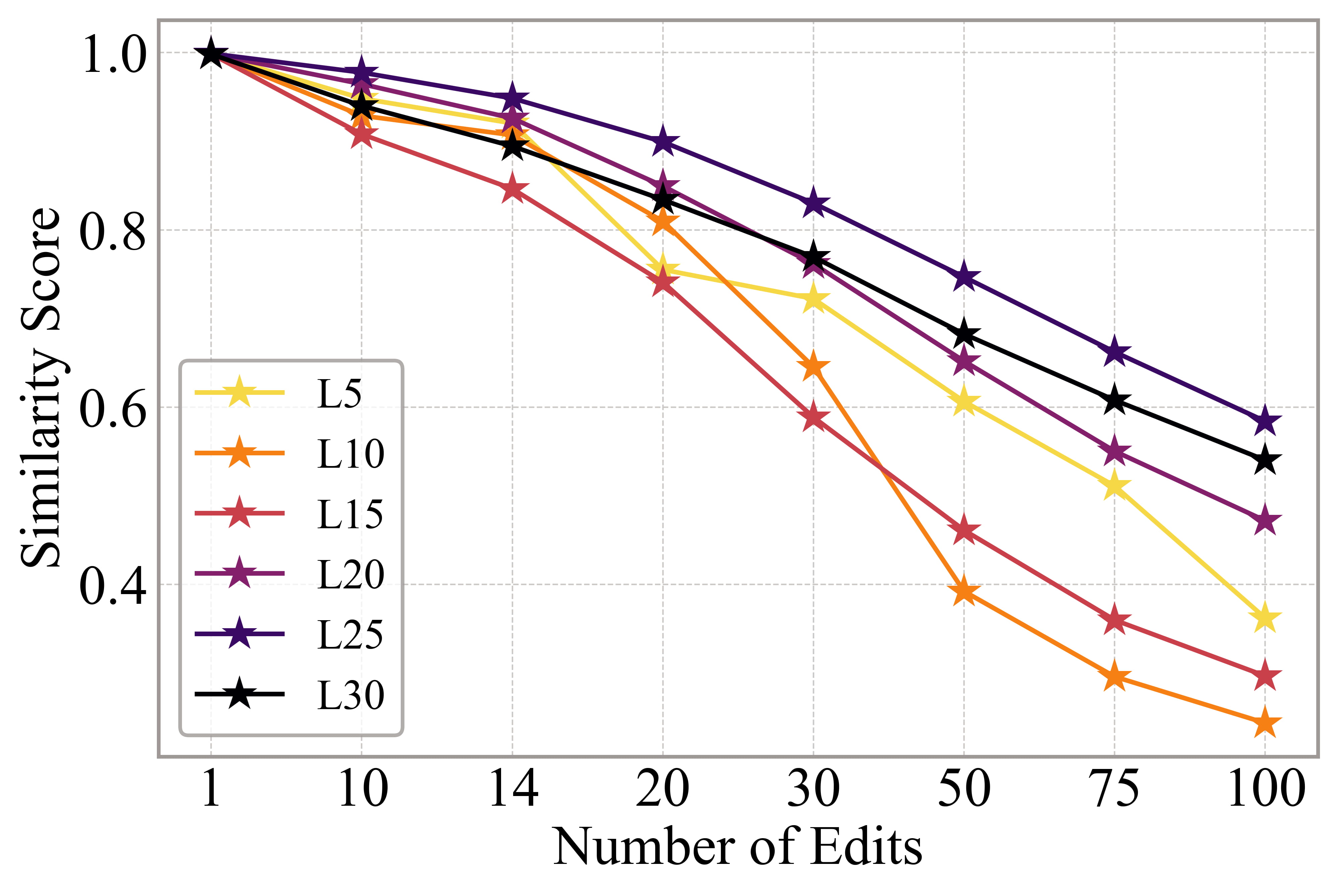}
  \vspace{-3mm}
  \caption{Similarity score based on the Pearson product-moment correlation coefficient, calculated between the parameters of the original and edited model layers.}
  \vspace{-4mm}
  \label{Fig:similarity-score}
\end{figure}

As illustrated in Figure \ref{Fig:similarity-score}, with fewer than 15 edits, the correlation coefficient (\(R\)) between the edited and original layers remains high (e.g. close to 1), indicating the significant similarity of the parameters. However, with an increasing number of edits, there is a marked decrease in similarity. Such changes in the parameter lead to a ``mismatch'' between the edited and original layers, which undermines the model's inherent coherence through layers. Consequently, the model's general capabilities are significantly damaged.

One interesting finding is that modifications in deeper layers, especially the 20th, 25th, and 30th layers, maintain relatively higher similarity scores compared to editing the shallower layers. This finding aligns with the experiments of editing different layers in Section \ref{sec:impact-of-different-editing-settings}, where we find that editing deeper layers results in a less pronounced decrease in performance. This distinction highlights a key architectural characteristic of LLMs: deeper layers, located closer to the output, exhibit greater tolerance to modifications, effectively sustaining the model’s performance. On the other hand, the shallower layers, forming the foundational processing stages of the LLMs, are more susceptible to disruptions from edits, leading to more significant performance degradations. This layered sensitivity within LLMs underscores the importance of strategic layer selection in the editing process.

We argue that the diminishing similarity between the edited and original layers is a primary factor in the model's reduced performance, disrupting its internal coherence and substantially impacting its performance in various tasks.

\subsection{Language Modeling Capability}
\label{sec:language-modeling-capability}

We hypothesize that the significant changes in the edited layers damage the language modeling capability of LLMs. 
To validate this hypothesis, we use \texttt{Vicuna-7b-v1.5} \cite{zheng2023judging} to measure the \textit{Perplexity} (PPL) of output sequences generated by post-edited models edited by ROME. CommonsenseQA is used as the evaluation dataset.

In our setting, we concatenate each question with its corresponding generated answer and calculate the \textit{perplexity} solely for the first 20 tokens of the answer portion. Answers with less than 20 tokens are excluded to avoid the effect of sequence length on the PPL. 
Additionally, we observe that in certain instances, the post-edited models tend to produce repetitive token sequences, which, while contributing to lower perplexity scores, are not meaningful in the context of answering CommonsenseQA questions. To address this, we implement a penalty ratio for repetitive sentences to ensure a more accurate reflection of the model's language modeling capability. The details of the formula to calculate the adjusted perplexity are shown in Appendix \ref{appendix:language-modeling-capability}.

As illustrated in Table \ref{tab:perplexity-results}, after 100 sequential edits, editing the 10th and 15th layers results in an extremely high perplexity, which leads to a zero score for the performance. On the other hand, editing the 5th layer results in a relatively low perplexity, indicating that the model is not completely damaged, although there is a significant decline in the performance as shown in Table \ref{tab:eval-results}. Editing the 20th layer maintains a lower perplexity, which guarantees a high performance on CommonsenseQA. These findings can explain the observations in Figure \ref{Fig:compare-edit-layer}. However, although editing the 25th and 30th layers severely damages the language modeling capability of LLMs, they still maintain very high performance on CommonsenseQA, as shown in Figure \ref{Fig:compare-edit-layer}. We explain this by examining the in-context learning capability in Section~\ref{sec:in-context-learning-capability}.

\subsection{In-Context Learning Capability}
\label{sec:in-context-learning-capability}
We further investigate whether, after memory editing, LLMs can still maintain the in-context learning capabilities. \citet{wang2023label} demonstrates that in in-context learning, the shallow layers of LLMs aggregate information from contexts into label words (for example, the CommonsenseQA contains five options as label words - A, B, C, D, or E), while in deep layers, LLMs extract and use the aggregated information of label words to perform the final prediction.  Inspired by this work, we evaluate the post-edited LLM on SST2 \cite{socher2013recursive} where the label words are ``positive'' and ``negative'', based on 1-shot in-context learning. We use \texttt{LLaMA-2-7B} as the base LLM and edit it using ROME and MEMIT on different layers. To save space, we describe the experimental setup and detailed results in Appendix \ref{appendix:in-context-learning-capability}. The experimental results indicate that editing shallow (e.g. the 5th layer) and deep layers (e.g. 20th, 25th, and 30th layers) does not significantly influence the in-context learning capability of LLMs. 

These findings also explain the phenomenon mentioned in Section~\ref{sec:language-modeling-capability} -- although editing the 25th and 30th layers severely damages the language modeling capability of LLMs, they still maintain very high performance on CommonsenseQA as shown in Figure \ref{Fig:compare-edit-layer}. The experiments illustrated in Figure \ref{Fig:compare-edit-layer} on CommonsenseQA are based on 8-shot in-context setting, and the first token of the generated sequence is treated as the final prediction. Given the maintenance of in-context learning capability, the post-edited model is still able to correctly predict the first token of the generated sequence, although it fails to generate a meaningful sentence because of the damage to language modeling capability.

\section{Conclusions}

We conduct a comprehensive evaluation of two types of memory editing methods for LLMs across eight diverse benchmarks. Our findings indicate that parameter-modifying ME methods tend to systematically degrade the model performance on general downstream tasks. In contrast, the parameter-preserving ME method, GRACE, successfully maintains the LLMs' capabilities but fails to maintain \textit{generalization}. We also show that increasing model size, instruction tuning, editing deeper layers, and increasing the batch size of memory editing are beneficial to mitigate the damage of parameter-modifying ME methods to LLMs. Finally, we conduct an in-depth analysis of how parameter-modifying ME methods hurt the general capabilities of LLMs. Overall, our research provides comprehensive insights into the dynamics of how, when, and why memory editing influences LLMs, offering valuable guidance for future research on memory editing.

\section{Limitations}

Despite the contributions, our study still has limitations. Our experiments on parameter-preserving ME methods are not exhaustive. As shown in Figure \ref{Fig:compare-edit-layer}, there is an observed performance decrease after 100 edits when editing layers 20/25 with ROME. Further experiments are needed to understand these long-term effects. Besides, we do not completely explain why LLMs can maintain in-context learning capabilities after being sequentially edited. These limitations highlight areas for future research, underscoring the need for more extensive investigations to refine our understanding of the intricate balance between knowledge editing and model integrity in LLMs.

\bibliography{custom}

\appendix

\newpage

\section{Editing Methods}
\label{sec:appendix-ediiting-methods}

\begin{table*}[hbpt]
\centering
\small
\begin{tabular}{l|ccccc}
\toprule
 & Method  & \makecell{Additional \\Training}  & Edit Layer & Default Edit Parameter  \\
\midrule Preserving Parameters & GRACE  & NO & FFN & $30^{th} mlp_{proj}$ \\

\midrule \multirow{3}{*}{Modifying Parameters} & MEND  & YES & FFN & $Model_{hyper} + 29/30/31^{th} mlp$  \\
 & ROME  & NO & FFN & $5^{th}\ mlp_{proj}$ \\
 & MEMIT  & NO & FFN & $4/5/6/7/8^{th}\ mlp_{proj}$ \\
\bottomrule
\end{tabular}
\caption{The details of memory editing methods. The edit parameter is in default for all checkpoints. We also conduct the ablation study on edited layers where we specify the exact layers we edit. In the table, $mlp_{proj}$ means the down project layer of the MLP layer, while $mlp$ means we edit the gate/up/down project layers of the MLP layer.}
\label{tab:summary-methods}
\end{table*}

We conduct our experiments on four ME methods. The summary of each method is shown in Table \ref{tab:summary-methods}. We introduce GRACE and ROME in detail in the following sections. The MEMIT method is not introduced as it is an improved version of ROME.

\subsection{GRACE}
\label{appendix:grace}

GRACE \cite{hartvigsen2022aging} is a method designed for sequential memory editing without altering original model parameters. The GRACE adapter, which is wrapped into a chosen layer of an LLM, contains two components: (1) a codebook that consists of a set of keys, denoted as $\mathbb{K}$, and values, denoted as $\mathbb{V}$, and (2) deferral radii, denoted as $\mathcal{E}$, to decide whether the input information flow uses the codebook. Specifically, $\mathbb{K}$ is a set of cached activation $h^{l-1}$ predicted by layer $l-1$. $\mathbb{V}$ is a set of values that are randomly initialized and updated using the LLMs' loss for edits. Each key is corresponding to a single value. The hyperparameter $\epsilon \in \mathcal{E}$ is a threshold for the similarity between the new input and previous edited knowledge. GRACE adapter is activated at layer $l$ only if this similarity is smaller than the radius. 

During editing, GRACE adds keys, corresponding values, and $\epsilon$ entries. In the inference process, at layer $l$, if the similarity of the activation at layer $l-1$ and keys are smaller than the corresponding radius $\epsilon$, the activation of the next layer becomes the cached corresponding values. Formally, the activation of $l$th layer is formulated as follows:

\begin{equation}
h^l= \begin{cases}\operatorname{GRACE}\left(h^{l-1}\right) & \text { if } \min _i\left(d\left(h^{l-1}, \mathbb{K}_i\right)\right)<\epsilon_{i_*}, \\ f^l\left(h^{l-1}\right) & \text { otherwise }\end{cases}
\end{equation}
where $i_*=\operatorname{argmin}_i\left(d\left(h^{l-1}\right), \mathbb{K}_i\right)$ and $f^l\left(h^{l-1}\right)$ denotes the $l$-th layer's activation of the unedited model.  $\epsilon_i^l$ and $\mathbb{K}_i^l$ are the deferral radius and key $i$ in layer $l$. GRACE($h^{l-1}$) retrieves the corresponding value associated with the closest key. $d(.)$ is a distance function. Following \citet{hartvigsen2022aging}, we use Euclidean distance in our experiments.

As shown in our experiments in Section \ref{sec:trade-off-grace-threshold}, the hyperparameter $\epsilon$ is a trade-off between \textit{generalization} and maintaining the broader fundamental capabilities of LLMs.

\subsection{ROME and MEMIT}
\label{appendix:rome}

ROME \cite{meng2022locating} applies a Locate-then-Edit strategy, which first utilizes the causal tracing method to ensure that MLP layers in LLMs play a role in recalling factual knowledge, and then edits specific MLP layers to integrate new knowledge into LLMs. Following \citet{meng2022locating}, we denote the first layer of the $l$th MLP layer as $W_{fc}^{(l)}$, and the second layer as $W_{proj}^{(l)}$. ROME treats $W_{proj}^{(l)}$ as a linear associative memory, which claims that any linear operation $W$ can work as a key-value store for a set of Key-Value vectors denoted as $K=\left[k_1\left|k_2\right| \ldots\right]$ and $V=\left[v_1\left|v_2\right| \ldots\right]$, respectively. A new key-value pair $(k_*, v_*)$ can be injected into $W$ by solving the following equation:
\begin{equation}
\operatorname{minimize}\|\hat{W} K-V\| \text { such that } \hat{W} k_*=v_*.
\end{equation}

This can be solved by setting $\hat{W}=W+\Lambda\left(C^{-1} k_*\right)^T$, where $W$ is the original matrix, $C=KK^T$ is a pre-cached constant, and $\Lambda=\left(v_*-W k_*\right) /\left(C^{-1} k_*\right)^T k_*$. In ROME's work, $C$ works as a constraint method to avoid edited parameters forgetting other unrelated knowledge. It is computed using the hidden states $k$ of 100,000 random samples from Wikipedia text. We evaluate whether the constraint method is beneficial to mitigating the damage of ROME to the general capabilities of LLMs in the Appendix \ref{appendix:efficacy-of-constraint-method-in-rome}.

MEMIT \cite{meng2022memit}, which can edit multiple knowledge at a time (e.g. batch editing), is a following work of the ROME \cite{meng2022locating}. 

\section{Evaluation Datasets}
\label{sec:appendix-evaluationg-datasets}

To rigorously assess the impact of ME methods on LLMs, we employ a diverse set of benchmarks encompassing essential capabilities, including Professional Knowledge, Common Sense Knowledge, Logical Reasoning, Reading Understanding, and Multilingual Proficiency. Our evaluation consists of eight benchmarks, the specifics of which are delineated in Table \ref{tab:benchmarks}. We leverage the \textit{opencompass} codebase \cite{2023opencompass}, a widely recognized open-source repository for LLMs evaluation. In alignment with their established protocols, we adopt the Perplexity (PPL) mode for the evaluation of the MMLU dataset. For instance, in the MMLU dataset, each item comprises a question and four possible answers. We concatenate the question with each answer option to create four distinct input sequences. 
Subsequently, we compute the Perplexity for each sequence using the edited LLMs under examination. A lower Perplexity score indicates higher model confidence in the corresponding sentence, thereby guiding our selection of the answer with the lowest score as the definitive prediction. Conversely, for the remaining benchmarks, we utilize the Generation (GEN) mode for evaluation. Specifically, for MATH, BBH, and TyDiQA, we ascertain the accuracy of the model's predictions against the ground truth following a post-processing procedure. Regarding the programming task MBPP, we employ Python's built-in \textit{exec()} function to verify the error-free execution of the generated code.

\begin{table*}[hbpt]
\centering
\small
\begin{tabular}{c|ccccccc}
\toprule
Capability & Task & Datasets & \#. Items & Metrics & Language & Mode & \#. Shots\\
\midrule \makecell{Professional\\Knowledge} & \makecell{High School / University \\ Professional Examination} & MMLU & 15691 & Acc. & English & PPL & 5\\
\midrule \multirow{3}{*}{ \makecell{Logical\\Reasoning} } & Mathematical Reasoning & MATH & 5000 & Acc. & English & GEN& 4\\
 & Comprehensive Reasoning & BBH & 6511 & Acc. & English & GEN & 3\\
 & Textual Entailment & AX-b & 1104 & Acc. & English & GEN & 0\\
\midrule \makecell{Common Sense\\Knowledge} & \makecell{Knowledge \\Question Answering} & ComQA & 1221 & Acc. & English & GEN & 8\\
\midrule \makecell{Reading\\Understanding} & Reading Understanding & C3 & 1825 & Acc. & Chinese & GEN & 0\\
\midrule \makecell{Multilingual\\Proficiency} & \makecell{Multi-Language \\ Question Answering} & TyDiQA & 6322 & F1 & 13 languages & GEN & 0\\
\midrule \makecell{Code\\Generation} & Code Generation & MBPP & 500 & Pass. & Code & GEN & 3\\
\bottomrule
\end{tabular}
\caption{The details of downstream evaluation benchmarks.}
\label{tab:benchmarks}
\end{table*}



\section{The Trade-off of the Threshold in GRACE}
\label{appendix:trade-off-of-the-threshold-in-grace}

As shown in Table \ref{tab:grace-threshold}, the \textit{generalization} increases rapidly when we increase the threshold from 1 to 20. However, the capabilities of multi-lingual understanding and code generation are completely damaged. One counter-intuitive finding is that the performance of the MMLU is not hugely influenced. We leave the explanation of this phenomenon as future work.

\begin{table}[htbp]
    \centering
    \small
    \setlength\tabcolsep{4pt}
    \begin{tabular}{c|cccc|cc}
        \toprule
        $\epsilon$ & MMLU & ComQA & TyDiQA & MBPP & Rel. & Gen. \\
        \midrule
        1 & 46.8 & 44.1 & 26.8 & 15.8 & 99 & 30.2 \\
        5 & 46.8 & 44.1 & 23.0 & 15.2 & 98 & 45.0 \\
        10 & 46.8 & 39.7 & 22.8 & 15.2 & 99 & 52.1 \\
        20 & 46.2 & 12.1 & 0.3 & 0 & 98 & 97.3 \\
        \bottomrule
    \end{tabular}
    \caption{The evaluation results across different thresholds of GRACE. We edit the 20th layer in this experiment, which is different in Table \ref{tab:eval-results} where we edit the 30th layer. We denote $\epsilon$ as the threshold. Rel. and Gen. are \textit{reliability} and \textit{generalization} respectively, which is evaluated on the editing dataset. }
    \label{tab:grace-threshold}
\end{table}

\section{Additional Impact of Different Editing Settings}

\subsection{Efficacy of Constraint Method in ROME}
\label{appendix:efficacy-of-constraint-method-in-rome}

In our examination of ROME's constraint methodologies, which incorporate 100,000 Wikidata entries to limit the influence of edits on unrelated information, we analyze a variant of ROME without constraints (ROME w/o C). Figure \ref{fig:compare-checkpoints} illustrates that applying constraints significantly enhances the model's performance in all datasets, validating the effectiveness of this strategy. In the absence of constraints, a marked deterioration in performance is observed, notably in benchmarks like TyDiQA, CommonsenseQA, and MBPP. This finding indicates that unconstrained parameter modification can severely impair the model's efficacy, while the application of constraints attenuates this negative impact. However, it's noteworthy that the effectiveness of these constraints begins to wane after approximately 20 edits. This observation highlights an emerging need for innovative constraint methodologies in parameter modification, particularly in the context of sequential memory editing. Developing more robust constraint mechanisms could be vital to maintaining model performance and integrity over a broader range of edits.

\subsection{Explanation of Editing Deeper Layers}
\label{sec:Explanation-of-editing-deeper--layers}

In this section, we explain the phenomenon when we edit the 30th layer of LLaMA-2-7B using ROME. As shown in Figure \ref{Fig:compare-edit-layer}, after 14 edits to the 30th layer, the model's performance intriguingly plummeted to zero. However, a notable recovery occurred after 20 edits, with performance gradually increasing to approximately 45\% following 50 edits. This unusual pattern can be attributed to the methodology used in our evaluation, where we considered the first token of the output generated by the edited model as the final prediction. Initially, after 14 edits, the model's language modeling capability appeared to be completely compromised. Yet, after 20 edits, the model consistently predicted the first token as one of the candidates - 'A, B, C, D, or E' - although it still failed to generate a coherent sequence beyond this. This indicates that while the model retained the capacity to predict the first token accurately, its broader language modeling capabilities were significantly diminished. We delve into a more in-depth analysis and explanation of this phenomenon in Section \ref{sec:interpreting-memory-editing-disruptions-in-llms}, exploring this observation's underlying mechanisms and implications.

\subsection{Layers to Edit in GRACE Method}
\label{appendix:layers-to-edit-in-GRACE-method}

We also conduct experiments to edit different layers of LLaMA-2-7B using the GRACE method. According to Table \ref{tab:grace-layer}, with the same threshold, editing the shallower layer results in more damage to LLMs. This is because, in the shallow layer, the activations are not much different for different inputs because of the less calculation compared to deeper layers. We claim that editing deeper layers in GRACE is a better choice than that of shallower layers.

\begin{table}[htbp]
    \centering
    \small
    \setlength\tabcolsep{4pt}
    \begin{tabular}{c|cccc}
        \toprule
        Layer & MMLU & ComQA & TyDiQA & MBPP\\
        \midrule
        10 & 23.1 & 8.7 & 0.1 & 0 \\
        20 & 46.8 & 39.7 & 22.8 & 16.4 \\
        30 & 46.8 & 46.4 & 23.42 & 17.4 \\
        \bottomrule
    \end{tabular}
    \caption{The evaluation results across different editing layers of GRACE. The threshold is set by 10.}
    \label{tab:grace-layer}
\end{table}

\begin{figure}[ht]
  \centering
  \includegraphics[width=\linewidth]{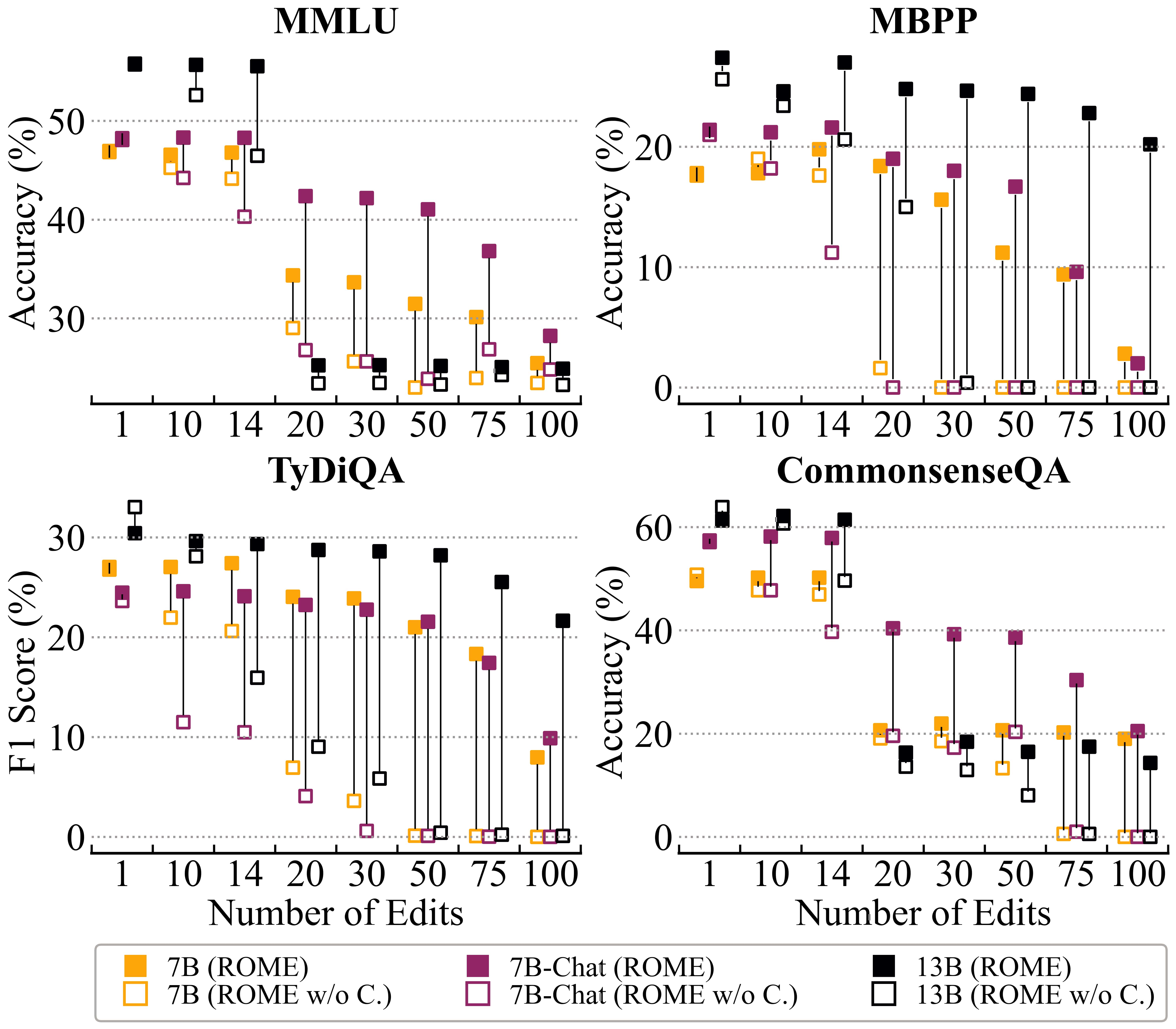}
  \caption{Evaluation Performance across three different checkpoints of LLaMA-2-7B. We denote the ROME method without constraint strategy using 100,000 Wikipedia text as  ROME w/o C.}
\end{figure}

\section{Additional Analysis of the Damage to LLMs by ME Methods}

\subsection{The Language Modeling Capability}
\label{appendix:language-modeling-capability}

As described in Section \ref{sec:language-modeling-capability}, we proposed adjusted perplexity as a measurement for the language modeling capability of post-edited LLMs to avoid the influence of the generated repetitive sequences. We employ Vicuna-7b-v1.5 \cite{zheng2023judging} to measure the Perplexity of the output sequences generated by post-edited models edited by ROME to answer questions in the CommonsenseQA dataset. Specifically, denote a generated sequence with $n$ tokens as $Y = (y_1, y_2, ..., y_n)$, we calculate the perplexity using the following equation:
\begin{equation}
\operatorname{PPL}(Y)=\exp \left\{-\frac{1}{t} \sum_i^t \log p_\theta\left(y_i \mid y_{<i}\right)\right\}
\end{equation}
where $p_{\theta}(y_i|y_{<i})$ is the log-likelihood of the $i$th token conditioned on the previous tokens $y_{<i}$. However, such a naive approach is not applicable in our situation because post-edited models tend to generate repetitive tokens, which leads to relatively low perplexity. Therefore, we calculate the n-gram repetitive ratio for each sequence. We first slice the sequence into several n-gram fragments, then we set the ratio of the number of unique fragments over the total number of fragments as the repetitive ratio $\rho$. Finally, we calculate the adjusted PPL is calculated by:
\begin{equation}
\operatorname{Adj\_PPL}(Y)=\operatorname{PPL}(Y) \times e^{1-\rho}
\end{equation}

\subsection{The In-Context Learning Capability}
\label{appendix:in-context-learning-capability}

In-context learning, which concatenates several demonstration-label pairs and the demonstration to be predicted as input context, is one of the most important capabilities of LLMs. \citet{wang2023label} explain the success of LLMs in in-context learning, that in the shallow layers (near to input), the model aggregates information from demonstrations to label words, while in deep layers, the model extracts and uses this information from previous label words to form the final prediction. In this section, we utilize the same way proposed by \citet{wang2023label} to analyze whether the in-context learning capability has been influenced after sequential edits. Specifically, we calculate the saliency score \cite{simonyan2013deep} for each attention matrix:

\begin{equation}
I_l=\left|\sum_h A_{h, l} \odot \frac{\partial \mathcal{L}(x)}{\partial A_{h, l}}\right|
\end{equation}
where $\mathcal{L}(x)$ is the loss function of the task, $A_{h,l}$ represents the value of attention matrix of the $h$-th attention head in the $l$-layer and $x$ represents the input. $I_l(i,j)$ is the significance of the information flow from the $i$-th token to $j$-th token. We denote $p_i$ as the $i$-th label words such as "True" or "False", $q$ as the target position in which the model predicts labels and $w$ as the words in demonstrations. $C$ represents the number of label words. We have three metrics as shown below:

$S_{wp}$: the saliency score of information flow from text part $w$ to label words $p$:

\begin{equation}
\begin{aligned}
S_{w p} & =\frac{\sum_{(i, j) \in C_{w p}} I_l(i, j)}{\left|C_{w p}\right|}, \\
C_{w p} & =\left\{\left(p_k, j\right): k \in[1, C], j<p_k\right\} .
\end{aligned}
\end{equation}

$S_{pq}$: the saliency score of information flow from label words $p$ to target position $q$:

\begin{equation}
\begin{aligned}
S_{p q} & =\frac{\sum_{(i, j) \in C_{p q}} I_l(i, j)}{\left|C_{p q}\right|}, \\
C_{p q} & =\left\{\left(q, p_k\right): k \in[1, C]\right\} .
\end{aligned}
\end{equation}

$S_{pq}$: the saliency of information flow except $S_{wp}$ and $S_{pq}$:

\begin{equation}
\begin{aligned}
S_{w w} & =\frac{\sum_{(i, j) \in C_{w w}} I_l(i, j)}{\left|C_{w w}\right|} \\
C_{w w} & =\{(i, j): j<i\}-C_{w p}-C_{p q}
\end{aligned}
\end{equation}

\begin{figure*}[ht]
  \centering
  \includegraphics[width=\linewidth]{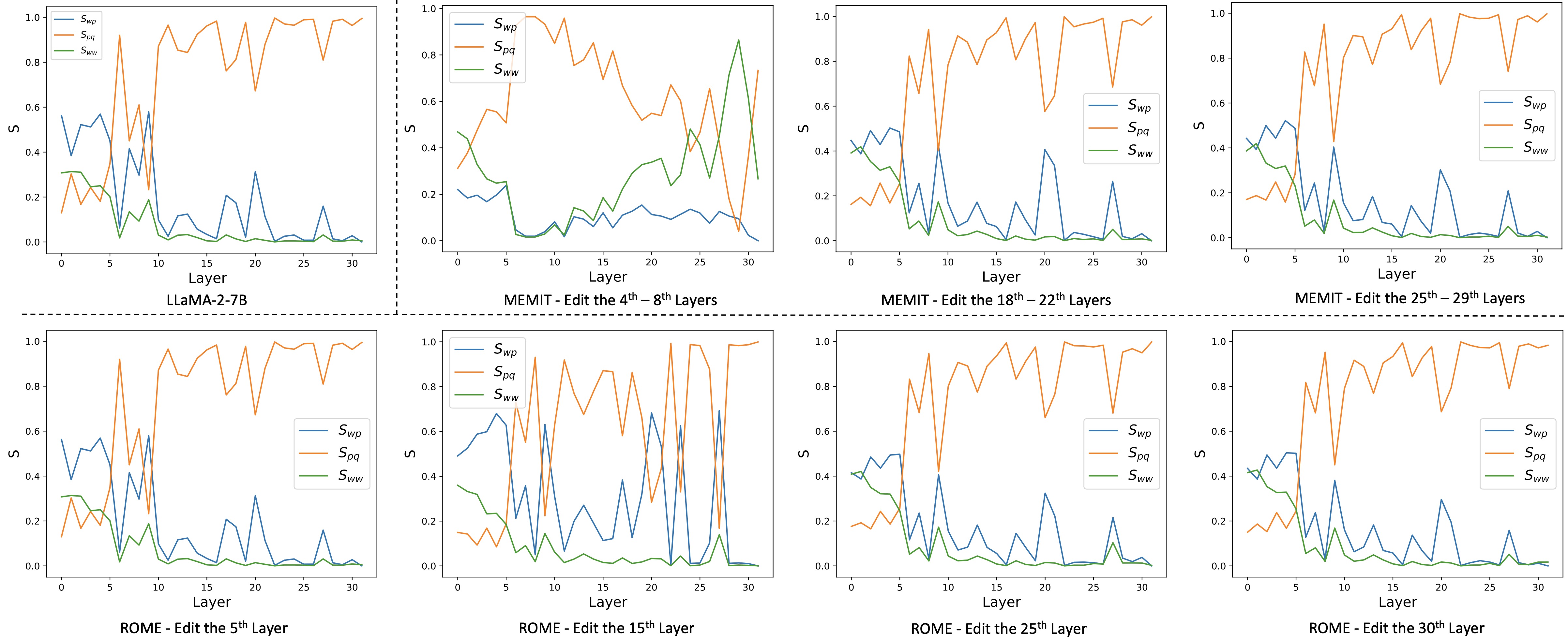}
  \caption{In-context learning saliency score 
 }
  \label{Fig:in-context-total}
\end{figure*}

We utilize SST-2 \cite{socher2013recursive} as the experimental datasets and one-shot setting. According to Figure \ref{Fig:in-context-total}, the original Llama-2-7B model proves the claim proposed by \citet{wang2023label}. Specifically, in the shallow layer (from layer 0 to layer 5), the line of $S_{wp}$ dominates, which shows that the information is aggregating from text to labels. While in the deep layer (from layer 6 to the last layer), the line of $S_{pq}$ dominates, indicating that the label information is aggregating to the target position. For the ROME method, editing layer 5 has a slight influence on layers 6 to 10, which promotes the information aggregating to label words process. Because the change is not very obvious, the model can still maintain an average score of 18.3\% accuracy according to Table \ref{tab:eval-results}. While if we edit layer 15, due to the damage stored in layer 15, in the deeper layer, there are some fluctuate between $S_{wp}$ and $S_{pq}$, which shows unstable attention across those layers, resulting in much worse performance on CommonsenseQA as shown in Figure \ref{Fig:compare-edit-layer}. The same thing happens when we edit layers from 4th to 8th using the MEMIT method. It is shown that in the deeper layer, the information fails to aggregate form label words to target position, which explains a worse average score of 3.8\% according to Table \ref{tab:eval-results}. Finally, editing the 30th layer does not have much influence on such attention mechanism for information flow. This means that the perplexity capability is much different from the in-context learning capability. Besides, this also partly explains why editing the 30th layer using ROME gives a high performance after 100 edits.

\end{document}